\documentclass[10pt,twocolumn,letterpaper]{article}

\usepackage{iccv}              

\definecolor{iccvblue}{rgb}{0.21,0.49,0.74}
\usepackage[pagebackref,breaklinks,colorlinks,allcolors=iccvblue]{hyperref}

\def\paperID{*****} 
\def\confName{ICCV}
\def\confYear{2025}

\title{Online 3D Multi-Camera Perception through Robust 2D Tracking and \\Depth-based Late Aggregation}

\author{Vu-Minh Le$^{1,2,\text{†},*}$ \quad Thao-Anh Tran$^{1,3,*}$ \quad Duc Huy Do$^{1,2,*}$ \quad Xuan Canh Do$^{2,*}$\vspace{0.15em}\\
Huong Ninh$^{1,2}$ \quad Hai Tran$^1$
\vspace{0.3em}\\
$^{1}$ Optoelectronics Center, Viettel Aerospace Institute, Viettel Group\\
$^{2}$ University of Engineering and Technology, Vietnam National University\\
$^{3}$ School of Electrical and Electronic Engineering, Hanoi University of Science and Technology
}

\begin{document}
\maketitle
\begin{abstract}
Multi-Target Multi-Camera Tracking (MTMC) is an essential computer vision task for automating large-scale surveillance. With camera calibration and depth information, the targets in the scene can be projected into 3D space, offering unparalleled levels of automatic perception of a 3D environment. However, tracking in the 3D space requires replacing all 2D tracking components from the ground up, which may be infeasible for existing MTMC systems. In this paper, we present an approach for extending any online 2D multi-camera tracking system into 3D space by utilizing depth information to reconstruct a target in point-cloud space, and recovering its 3D box through clustering and yaw refinement following tracking. We also introduced an enhanced online data association mechanism that leverages the target's local ID consistency to assign global IDs across frames. The proposed framework is evaluated on the 2025 AI City Challenge's 3D MTMC dataset, achieving 3rd place on the leaderboard.

\end{abstract}    
\section{Introduction}
\label{sec:intro}

\let\thefootnote\relax\footnotetext{* Equal contributions.}
\let\thefootnote\relax\footnotetext{$\text{†}$ Corresponding author's email: \href{mailto:lvm.lvm.2003@gmail.com}{lvm.lvm.2003@gmail.com}}

Multi-Target Multi-Camera Tracking (MTMC), the task of detecting all objects on multiple cameras and assigning consistent IDs to unique entities in the scene, is a crucial computer vision problem for automatic processing of the large amount of data that surveillance cameras produce. In a multi-camera environment, with proper camera calibration and depth information, 3D bounding boxes are excellent representations of the target's location, providing great detail for 3D scene perception.

While conventional MTMC methods have achieved remarkable performance in 2D space \cite{xie2024robust, yoshida2024overlap, kim2024cluster}, extending this capability to 3D remains a challenging problem. The primary obstacles arise from the need to effectively fuse multi-view information, handle partial occlusions, and represent object-level 3D geometry in a scalable and interpretable way. To resolve this, recent approaches such as EarlyBird and BEV-SUSHI \cite{teepe2024earlybird, wang2024bev} opt to fuse multi-view image data or detections into a unified 3D space before tracking. While these methods have been shown to achieve state-of-the-art performance on 3D MTMC tasks, the task of transitioning established 2D MTMC systems to 3D tracking typically requires replacing all 2D components, which is impractical for large-scale surveillance systems due to the complexity and cost of such overhauls.

\begin{figure}
\begin{center}
\includegraphics[width=\linewidth]{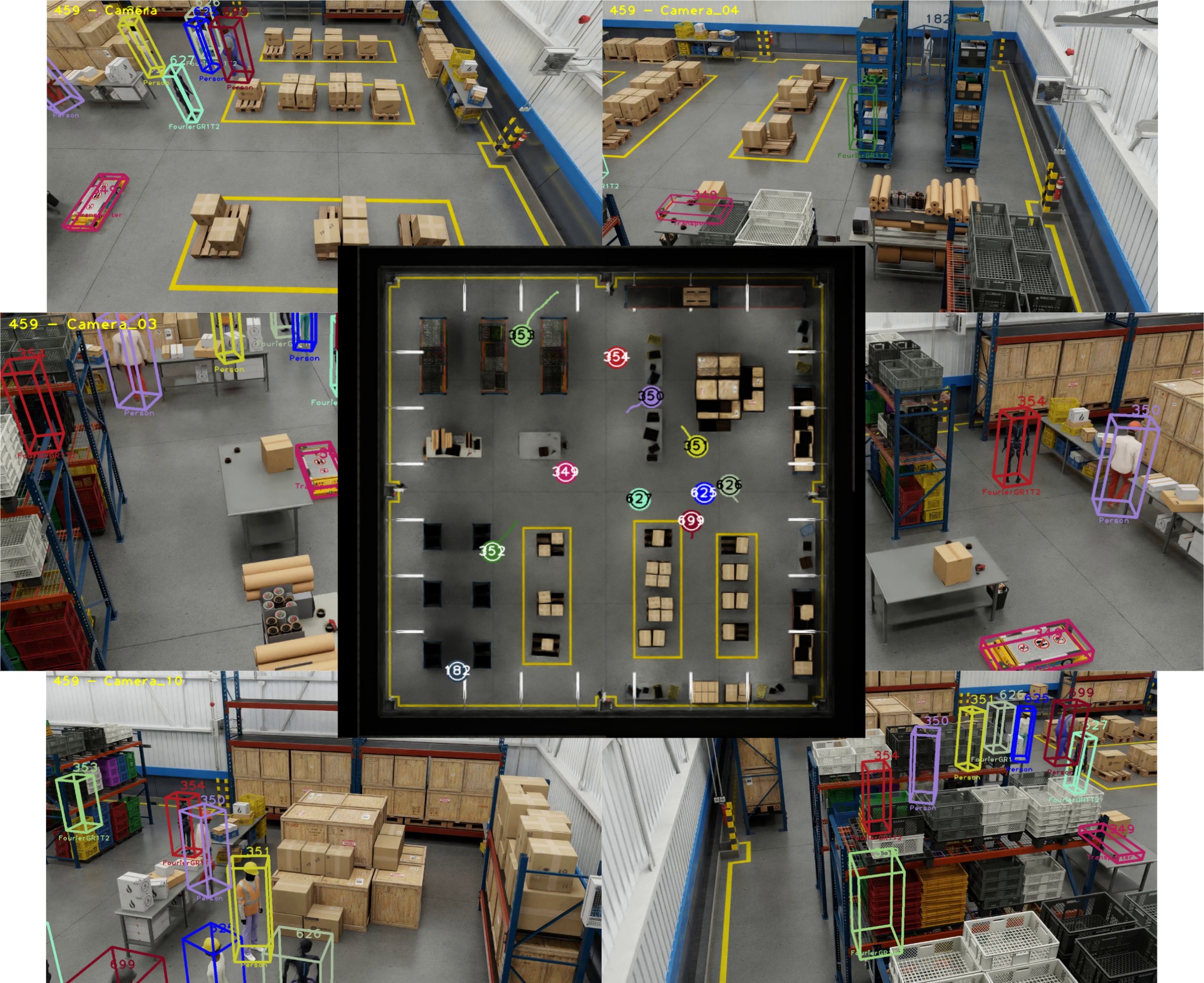}
\end{center}
\vspace{-1em}
   \caption{A depiction of the 3D Multi-Camera Tracking problem, where a target in the space is represented by a single 3D bounding box, and can be projected to the 2D view of all cameras in the scene. The task is to localize the 3D bounding boxes of these targets and assign them consistent IDs as they traverse across the scene.} 
\label{fig:mtmc_3d_demo}
\end{figure}

To address this problem, this paper proposes a novel framework that enables Online 3D Multi-Target Multi-Camera Tracking by constructing the target in the point-cloud space with the aid of the 2D MTMC results, then recovering the 3D bounding box in the space. Our framework consists of two main stages. First, we perform 2D multi-camera tracking to obtain 2D bounding boxes and consistent identities across views, where we introduce an enhanced association mechanism for assigning global IDs to targets using their local ID consistency. Then, we fuse depth information from multiple cameras to reconstruct localized point clouds per tracked object. These point clouds are transformed into a global coordinate system using extrinsic calibration parameters, creating a 3D representation of the object. A bounding box around the object is then created and refined to yield the final tracking result.

Our contributions in this paper are as follows:

\begin{itemize}
    \item We propose a framework that flexibly extends the traditional online 2D MTMC systems to output 3D tracking results through late aggregation. 
    \item We introduce an enhanced online association mechanism that improves ID assignment performance on cross-camera tracking by utilizing each target's local ID as a feature to propagate the global ID across frames.
    \item We present a mechanism for extending the ID-assigned 2D bounding boxes into the 3D space by leveraging segmentation and point-cloud clustering on the depth maps to create 3D bounding boxes.
    \item Our proposed methods are validated on the 2025 AI City Challenge 3D MTMC dataset \cite{Tang25AICity25}, achieving third place on the track's public leaderboard.
\end{itemize}

\section{Related Works}
\label{sec:related-works}

\subsection{Object Detection}

Object detection aims to localize and classify all relevant objects in RGB images, which is a crucial step in the majority of Multi-Target Tracking algorithms. Recent methods fall primarily into two categories: CNN-based and Transformer-based approaches. CNN-based methods are further divided into anchor-based and anchor-free approaches. Anchor-based detectors \cite{fastrcnn,fasterrcnn,maskrcnn,cascadercnn} rely on predefined bounding boxes (anchors) for predicting object locations, achieving high accuracy, although at the expense of flexibility for objects of unconventional scales and aspect ratios. Anchor-free CNN methods \cite{yolov8,yoloX,yolov11,objectsaspoints} predict object locations directly without anchors, demonstrating improved real-time efficiency and robustness across diverse scales. Transformer-based detectors \cite{detr,deformable,dino,rtdetr,ViT-Det, zong2023detrs} have emerged as a promising upgrade, employing self-attention mechanisms that effectively capture long-range contextual dependencies, leading to enhanced localization and classification performance, especially on small and similar objects, albeit requiring extra computational overhead. 

\subsection{Multi-Target Multi-Camera Tracking} 

In recent years, Multi-Camera Tracking has seen great developments. Most 2D approaches typically start by tracking on individual cameras first, before aggregating the results from all cameras and synchronizing the IDs for the same entity in the scene \cite{iguernaissi2019people}. During single-camera tracking, the generated bounding boxes from the object detectors first have their features extracted, such as appearance features \cite{fairmot, li2023clip, wojke2017simple}, motion \cite{ocsort, maggiolino2023deep, bewley2016simple}, and pose information \cite{kim2024cluster, xie2024robust, yoshida2024overlap}. The extracted information is then used to facilitate the association of the bounding boxes between multiple entities in different frames using various heuristics \cite{bewley2016simple, wojke2017simple, bytetrack, tran2024advancing, ocsort, maggiolino2023deep} or learnable methods \cite{wang2024bev, nguyen2022lmgp, cheng2023rest, cameltrack}. These methods are categorized into the tracking-by-detection paradigm.

Afterwards, Multi-Camera Tracking makes use of the extracted appearance features to synchronize the IDs between cameras for all unique targets in the scene \cite{yao2022city, nguyen2023multi, zhang2017multi, huang2023enhancing, specker2022toward}. In a multi-view setting, camera calibration can be leveraged to further enhance association accuracy, through homography matrices \cite{kim2024cluster, cherdchusakulchai2024online, specker2024ocmctrack, yang2024online}, or epipolar geometric affinity \cite{xie2024robust}. The cross-camera matching problem is then solved through various optimization frameworks, with the two most common being view-by-view Hungarian matching \cite{kim2024cluster, huang2023enhancing} and hierarchical clustering \cite{nguyen2023multi, specker2024ocmctrack, cherdchusakulchai2024online}.

Online approaches pose the additional issue of ID assignment and management, as the algorithm may only utilize information from past frames to assign IDs to incoming detections, as opposed to the full trajectories that offline methods have access to. Recent approaches utilize the Hungarian algorithm to match clusters to existing tracks based on appearance and spatial features \cite{kim2024cluster, kim2023addressing}, doing away with the one-to-many mapping constraint between MOT and MTMC tracks entirely. Several works propose the idea of splitting the track based on appearance features \cite{specker2024ocmctrack, yang2024online} or delaying the assignment of ID \cite{specker2022toward} to prevent error accumulation. 

\begin{figure*}[!ht]
\begin{center}
\includegraphics[width=0.9\linewidth]{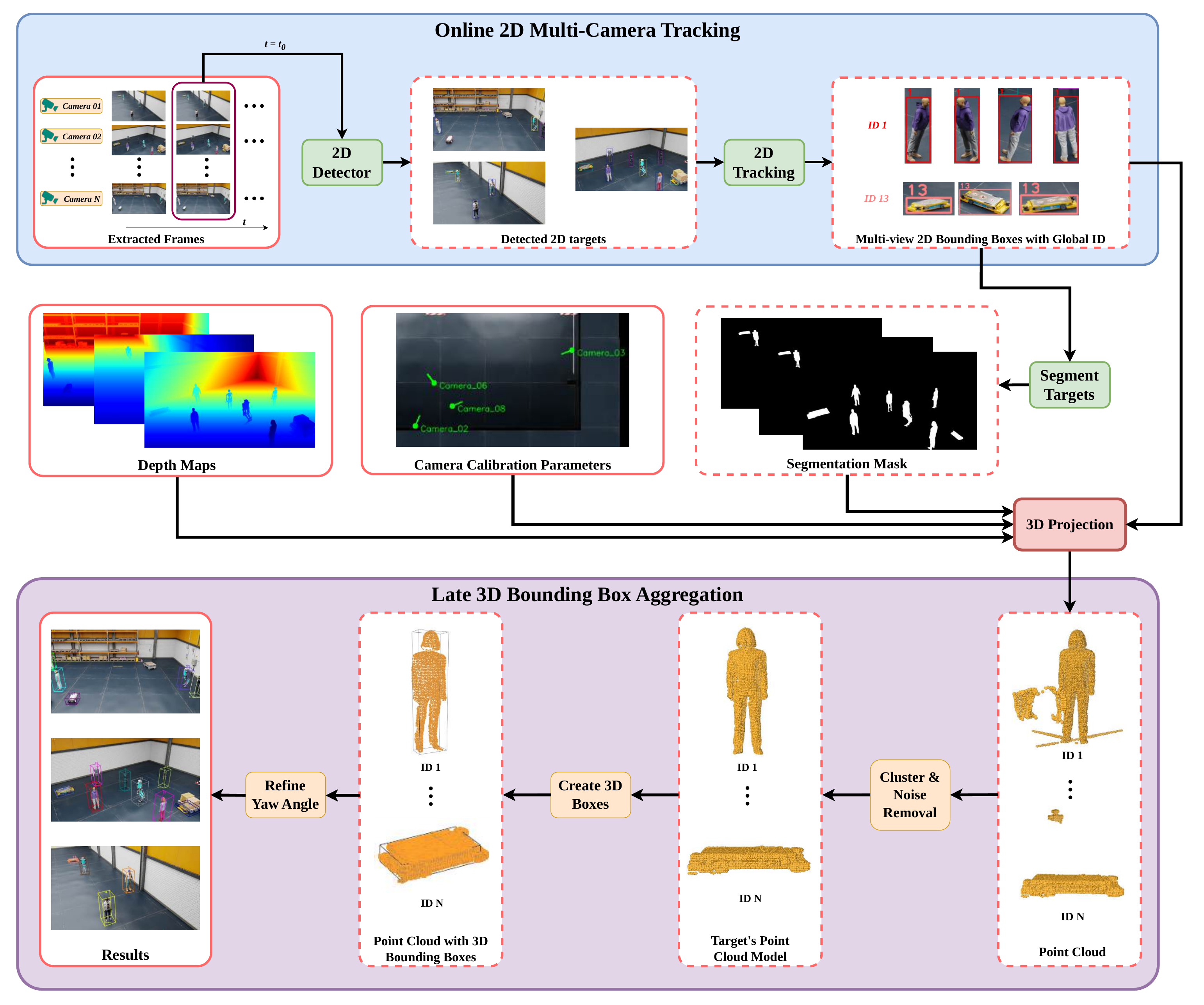}
\end{center}
\vspace{-1em}
   \caption{An illustration of our Online 3D Multi-Target Multi-Camera Tracking pipeline with Late 3D Bounding Box Aggregation.} 
\label{fig:mtmc_3d_pipeline}
\end{figure*}

\subsection{2D-Fusioned 3D Object Localization}
Estimating 3D positions from 2D images is essential when LiDAR is unavailable. Monocular methods \cite{disentangling,pseudolidar,smoke} infer 3D coordinates from RGB and geometric priors at low cost \cite{monopair,m3d,fcos3d}, but struggle with depth ambiguity in occluded or truncated scenes. Stereo methods \cite{chen2020dsgn,stereorcnn} improve depth via dual views but require precise calibration. Some models \cite{monodetr,gupnet} use contextual cues to enhance accuracy but degrade in sparse or complex scenes. Depth-distribution models \cite{caddn,deepstereo} increase precision at the cost of complexity. Most approaches rely on 2D proposals followed by 3D regression or triangulation \cite{monoflex,rtm3d}, which can propagate 2D errors.

Fusion-based methods \cite{monodetrnext,bevformer,liftsplatshoot} combine RGB features with BEV or contextual cues to improve robustness and approach LiDAR-level accuracy at lower cost, though generalization remains a challenge. LiDAR-based methods \cite{pointpillars,wang2021pointaugmenting, vora2020pointpainting} offer high precision from point clouds but require expensive sensors. Our camera-depth fusion method bridges this gap by integrating RGB images with pre-computed depth, balancing cost-efficiency and localization accuracy.
\section{Methods}
\label{sec:methods}

Figure \ref{fig:mtmc_3d_pipeline} depicts our proposed framework's pipeline, comprising two key phases: 2D Multi-Camera Tracking and Late 3D Bounding Box Aggregation. In the first phase, 2D object detection and tracking occur across synchronized camera views, producing frame-level object trajectories and 2D bounding boxes in each camera's coordinate space. The second phase constructs 3D object representations by integrating depth information, segmentation masks, and tracking IDs to generate per-target local point clouds. These are transformed into a global coordinate system using camera calibration data and aggregated into dense point clouds. An unsupervised clustering algorithm groups points into coherent object clusters, removing noise, and oriented 3D bounding boxes are fitted to each cluster via geometric estimation and yaw refinement, yielding compact 3D representations of the targets in the scene.

\subsection{Multi-Object Tracking}

The single-camera tracking module of our tracker follows the popular tracking-by-detection paradigm, where the tracker's input is the detections from different time steps, and the core objective is to associate these bounding boxes together and assign them appropriate IDs through the features of the bounding boxes.

\noindent\textbf{Object Detection.} The 2025 AI City Challenge MTMC dataset features multiple different object classes, though with significant imbalances, especially with the FourierGR1T2 and AgilityDigit robotic classes making up less than $0.001\%$ of the 2D bounding boxes of the dataset. Thus, we leveraged the Transformer-based Co-DETR \cite{zong2023detrs} object detection model to detect targets in the scene. We also sampled the single training scene with the two classes more aggressively to mitigate this imbalance for training the detection model. 

\noindent\textbf{Target Feature Extraction.} For each detected target in the frame, we extract an appearance embedding vector and the pose information of the pedestrians, which are used as features to facilitate association in later steps. For ReID features, we utilize CLIP-ReID \cite{li2023clip} with the Vision Transformer image encoder to extract a 1280-dimensional vector, representing the object's appearance feature. For pose estimation, we directly used the RTMPose model \cite{jiang2023rtmpose}, pretrained on the CrowdPose dataset \cite{li2019crowdpose}, to infer 14 keypoints and their confidence scores on the detected persons in the frame. 

\noindent\textbf{Single-Camera Data Association.} As targets cross into each other's trajectories, ID switching and track loss become prevalent. To mitigate this, we employed the Deep OC-SORT Multi-Object Tracker for assigning IDs to the detected objects. Deep OC-SORT \cite{maggiolino2023deep} integrates OC-SORT \cite{ocsort} - an advanced approach to model the motion and recover the identification of targets through periods of occlusions and disappearance, with the Dynamic Appearance mechanism that considers the target confidence scores for establishing the association metric, and for updating the appearance vector representing the track.

\subsection{Enhanced 2D Multi-Camera Tracking}

Our Online Multi-Camera Tracking pipeline follows the decentralized tracking approach \cite{iguernaissi2019people, kim2023addressing, kim2024cluster}, where the detections assigned with local MOT IDs from all cameras at each time step are first associated together into clusters of the same entity (Spatial Data Association), then assigned global MTMC IDs based on the tracked clusters from past frames (Temporal Data Association). 

\subsubsection{Spatial Data Association}

At each timestep $t$, we aggregate all detected 2D targets in the scene and consider the extracted features for association. In this association step, we used two association metrics: ReID appearance features and the target's top-down location. The appearance features are extracted from the ReID model during the single-camera tracking stage, while the top-down location is obtained by selecting the 2D target's representative foot point and applying a projective transformation of the target location to the top-down map through the provided homography matrix $\mathbf{H}$. Pose estimation is utilized to extract the foot point for pedestrians with clearly visible legs. Otherwise, the bottom middle point is selected.

With the target features, distance matrices for each class are created for hierarchical target clustering. For pedestrians, the cosine distance between target ReID features is used, whereas the Euclidean distance between the target's top-down location is used for other classes, though all target pairs are subjected to maximum distance constraints of both features, along with the constraint that prevents two targets from the same camera from being clustered together. Finally, agglomerative clustering is applied to the distance matrices to create a set of clusters $\mathcal{G} = \{ \mathcal{G}_1^t, \mathcal{G}_2^t, ..., \mathcal{G}_N^t \}$ representing $N$ unique entities at time step $t$.

\subsubsection{MOT ID Consistency Temporal Association}

In temporal data association, we model the multi-camera track with 3 stages: Tentative, Confirmed, and Lost. Tentative tracks are newly created tracks in the system that require consecutive matching to be confirmed, while Lost tracks are tracks that have disappeared from all camera views. Clusters are matched with confirmed tracks first, then with lost and tentative tracks, respectively, before creating a new track ID. 

\noindent\textbf{MOT ID Consistency Matching.} To decide the ID to which the cluster $\mathcal{G}_i^t$ will be assigned, we introduce MOT ID Consistency, an approach for matching and performing track splitting in an online multi-camera tracking setting. The idea is that in consecutive frames, the clusters of the same target should maintain a consistent group of local target IDs. For example, at time step $t-1$, consider a multi-camera track $\mathcal{M}_i^{t-1}$, which maps to a set of single-camera (MOT) target IDs $\{ \mathcal{S}^{t-1}_a, \mathcal{S}^{t-1}_b, \mathcal{S}^{t-1}_c \}$ across 3 cameras. At time step $t$, it is expected that a cluster $\mathcal{G}_i^t$ with the same MOT ID grouping will emerge, or a subset of the group should the target exit a camera view. Thus, when assigning an ID to a cluster, we match its group of local IDs to that of the existing track, and assign global IDs to all targets in the cluster that were matched to the same ID in the previous time steps. We refer to these as ``expected" cluster elements, and are denoted as the MOT ID overlap between the MTMC track and the matched cluster $\mathcal{O}^t_i = \{ \mathcal{G}_i^t \cap \mathcal{M}^{t-1}_i \}$. Unmatched tracks after this stage are immediately set to Lost tracks.

\noindent\textbf{MOT ID Consistency Track Splitting.} However, problems such as ID switching or noisy location and appearance features during occlusion may lead to clusters with two or more different entities at a time step. During temporal association, cluster elements that have not been seen by the MTMC track, or ``unexpected" elements ${\mathcal{O}^*}^{t}_i$, defined by the complement of $\mathcal{G}^t_i$ in $\mathcal{M}^{t-1}_i$, or ${\mathcal{O}^*}^{t}_i =\{ \mathcal{G}^t_i \setminus \mathcal{M}^{t-1}_i \}$, are compared against the ``expected" elements $\mathcal{O}^t_i$, or the local MOT ID from the previous timestep competing for the same MTMC ID via the cosine similarity. 

\begin{figure}[t]
\begin{center}
\includegraphics[width=1\linewidth]{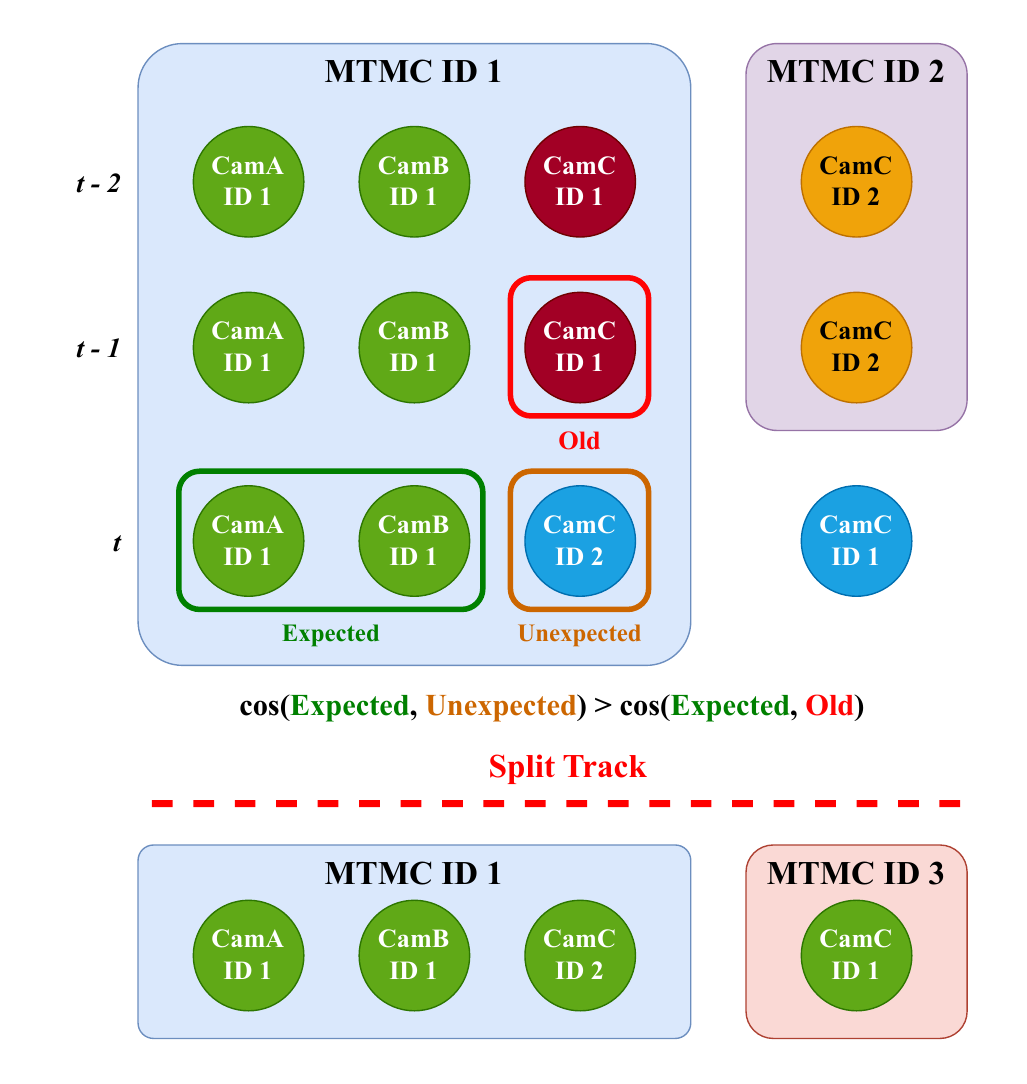}
\end{center}
\vspace{-1em}
   \caption{An example of our track splitting mechanism. At time step $t$, the track ID 2 from Camera C is correctly clustered back to the MTMC ID 1, and verified by a cosine similarity comparison with the ``expected" tracks. The falsely clustered ID 1 from Camera C in previous time steps is sent to be matched at later stages, eventually being assigned a new MTMC ID 3.}  
\label{fig:track_splitting}
\end{figure}

If the unexpected elements are more similar to the expected elements than the old local ID, we assume that the Spatial Data Association results from previous frames were faulty. Thus, we split the two MOT tracks, where from time step $t$ onwards, the more suitable MOT ID track is reassigned to the MTMC track. Otherwise, these ``unexpected" elements are sent to be matched in later association stages. Figure \ref{fig:track_splitting} provides a depiction of this process.

\noindent\textbf{Later matching stages.} Unmatched clusters are first paired with lost tracks via the Hungarian algorithm on ReID appearance embedding distances. A spatial constraint mandates that only cluster centroids within a radius of the lost track are admissible for rematching. This radius increases with the track's lost duration, though the number of consecutive matches a lost track must receive before reactivation also increases with it. Remaining clusters are then compared with tentative tracks under the same MOT ID Consistency paradigm as confirmed tracks. Tentative tracks become confirmed only after a fixed number of consecutive matches; otherwise, they are removed. Any clusters still unmatched beyond this point spawn new tentative tracks.

\subsection{Late 3D Bounding Box Aggregation}
\subsubsection{Background}
The dataset provides the 3D world coordinates $\mathcal W$ and $C$ camera coordinates $\{ \mathcal C_i \}^C_{i=1}$ corresponding to $C$ cameras. Additionally, each camera image is associated with a 2D image pixel coordinate system $\{ \mathcal P_i \}^C_{i=1}$. To project a point $\mathbf{X_c} = \{X_c, Y_c, Z_c\}^{\top}$ from $\mathcal C_i$ to a point $\mathbf{x} = \{ x, y \}^{\top}$ in $\mathcal P_i$, each camera $\mathcal C_i$ has an intrinsic matrix $\mathbf{K}_i$ defined as:

\begin{equation}
    \mathbf{K}_i = 
    \begin{bmatrix}
        f_{u_i} & 0 & c_{u_i} \\
        0 & f_{v_i} & c_{v_i} \\
        0 & 0 & 1
    \end{bmatrix}
\end{equation}
with $(c_{u_i}, c_{v_i})$ is the optical center and $(f_{u_i}, f_{v_i})$ is the focal length of the camera.

In addition, each camera is associated with an extrinsic matrix $\mathbf{E}_i = \left[ \mathbf{R}_i | \mathbf{t}_i \right]$ which projects a 3D point $\mathbf{X_w} = \{X_w, Y_w, Z_w\}^{\top}$ in $\mathcal W$ into $\mathcal C_i$. Here, $\mathbf{R}_i$ and $\mathbf{t}_i$ represent the rotation matrix and translation vector, respectively. To project a point from $\mathcal C_i$ back to $\mathcal W$, we use the inverse of the extrinsic matrix $\mathbf{E}_i^{-1} = \left[ \mathbf{R}_i^{\top} | - \mathbf{R}_i^{\top} \mathbf{t}_i\right]$.

\subsubsection{Depth to Point Cloud}
For each object, suppose there are $M$ ($M \le C$) 2D bounding boxes detected across $M$ cameras $\mathbf{\hat O_{2D}} = \{ \hat O_{2D_i} || \hat O_{2D_i} = (\hat B_{2D_i}, \hat s_{2D_i}, \hat c, \hat {id} )\}^M_{i=1}$ with the box location $\hat B_{2D_i}=( \hat x_{1_i}, \hat y_{1_i}, \hat x_{2_i}, \hat y_{2_i} )$, confidence score $\hat s_{2D_i}$, class ID $\hat c$ and object ID $\hat {id}$. For each 2D bounding box, we use SAM2 \cite{sam2} to segment the object inside the box, obtaining the mask of that object. This mask is then used to extract the depth $z_{k, i}$ of each $k$-th pixel belonging to the object from the depth map corresponding to each camera $i$. Then, we compute the camera coordinates $\mathbf{X_{c_{k, i}}} = \{X_{c_{k, i}}, Y_{c_{k, i}}, Z_{c_{k, i}}\}^{\top}$ of the point cloud, where each point corresponds to a pixel within the masked region on the depth map:

\begin{equation}
\begin{aligned}
    X_{c_{k, i}} &= \frac{(x_{k, i} - c_{u_i}) * z_{k, i}}{f_{u_i}} \\ 
    Y_{c_{k, i}} &= \frac{(y_{k, i} - c_{v_i}) * z_{k, i}}{f_{v_i}} \\ 
    Z_{c_{k, i}} &= z_{k, i}
\end{aligned}
\end{equation}

Then, all point clouds from the $M$ camera coordinate systems are projected into the world coordinate system:

\begin{equation}
\mathbf{X_{w_{k, i}}} = 
\begin{bmatrix}
X_{w_{k, i}} \\
Y_{w_{k, i}} \\
Z_{w_{k, i}}
\end{bmatrix}
= \mathbf{E}^{-1}_i
    \begin{bmatrix}
        \mathbf{X_{c_{k, i}}} \\
        1
    \end{bmatrix}
\end{equation}
Finally, we obtain a point cloud corresponding to the same object in the world coordinate system $\mathbf{P} = \{P_i || P_i = (\mathbf{X_{w_{k, i}}}, \hat s_{2D_i}, \hat c, \hat {id})\}_{i=1}^M$. 

\subsubsection{Point Cloud to 3D Bounding Box}

\begin{figure}[t]
\centering
\begin{subfigure}[b]{0.27\textwidth}
    \includegraphics[width=0.9\textwidth]{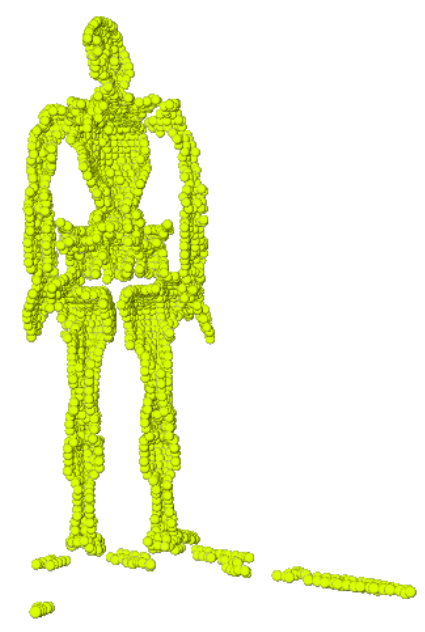}
    \centering
    \caption{Noise due to computational errors.}
    \label{fig:noise1}
\end{subfigure}
\begin{subfigure}[b]{0.13\textwidth}
    \includegraphics[width=0.9\textwidth]{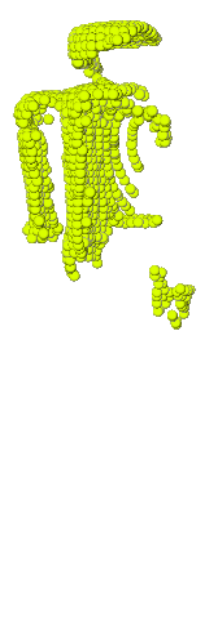}
    \centering
    \caption{Occluded parts.}
    \label{fig:noise2}
\end{subfigure}
\caption{(a) Noise from coordinate systems projection and points extraction from different cameras. (b) Multiple parts of the object are occluded and fragmented.}
\label{fig:noise}
\end{figure}

The resulting point cloud $\mathbf{P}$ often contains significant noise due to two main factors. First, computational errors can cause pixels of the same physical point from different cameras to appear slightly misaligned. Second, in crowded scenes, the 2D bounding box of an object may be occluded by other objects in some cameras. Thus, the reconstructed point cloud is often incomplete and does not fully capture the shape of the object. See Figure \ref{fig:noise} for more details.

To mitigate the impact of noise, we apply DBSCAN \cite{dbscan} to divide the point cloud into clusters based on point density. The algorithm takes two input parameters $epsilon$ is the maximum distance between two samples for them to be considered as in the same neighborhood and $min\_samples$ is the number of samples in a neighborhood for a point to be considered as a core point. We set $min\_samples = 50$ for all classes while $epsilon$ is chosen specifically for each class. A larger $epsilon$ is assigned to smaller-sized classes and a smaller $epsilon$ is used for larger-sized classes. This is because larger objects are generally less occluded, resulting in a denser point cloud, whereas smaller objects tend to produce a sparser and more fragmented point cloud. 

After clustering, we assume that the cluster containing the largest number of points corresponds to the object’s point cloud. We then compute the $xy$-center of the 3D bounding box by averaging the $x$ and $y$ of the points in the selected cluster. To estimate the box's length and width, we calculate the difference between the 95th and 5th percentiles of the $x$ and $y$ coordinates, respectively. The maximum $z$-coordinate within the cluster is used as the box height, and the $z\text{-center} = \text{height}/2$. However, we observe that if the clustered point cloud fails to form the actual shape of the object, the resulting bounding box may be smaller than the true size. Conversely, if the cluster contains a significant amount of noise, the predicted box may be excessively large. To address this, let $\hat V$ denote the volume of the predicted 3D bounding box, and let $V'_{\hat {c}}$ be the mean volume of class $\hat {c}$ in the training dataset, if $\hat V < \alpha_{lower} V'_{\hat {c}}$ or $\hat V > \alpha_{upper} V'_{\hat {c}}$, we set the predicted box dimensions are mean dimensions of class $\hat c$. We set $\alpha_{lower}=0.7$ and $\alpha_{upper}=1.5$ by default.

\subsubsection{3D Bounding Boxes Fusion}

\begin{figure}[t]
\centering
\begin{subfigure}[b]{0.22\textwidth}
    \includegraphics[width=1.1\textwidth]{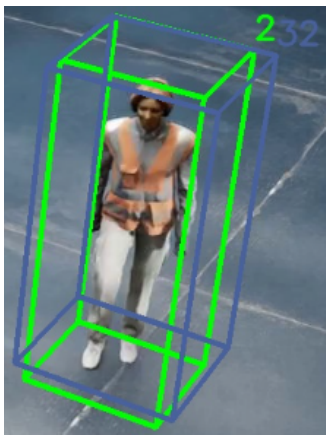}
    \centering
    \caption{Multiple 3D bounding boxes.}
    \label{fig:3dboxfusion1}
\end{subfigure}
\begin{subfigure}[b]{0.22\textwidth}
    \includegraphics[width=1.1\textwidth]{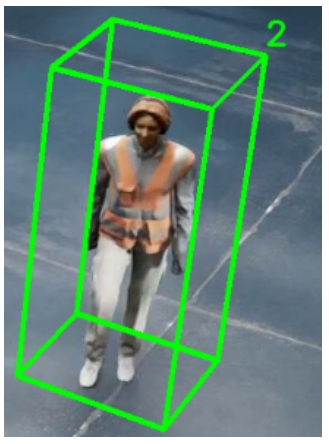}
    \centering
    \caption{Fused 3D bounding box.}
    \label{fig:3dboxfusion2}
\end{subfigure}
\caption{(a) A single object is assigned multiple 3D bounding boxes with different IDs. (b) The 3D bounding boxes are fused into a single 3D bounding box.}
\label{fig:3dboxfusion}
\end{figure}

As a result, each object being tracked is associated with a 3D bounding box constructed from its point cloud, forming a set of 3D bounding boxes $\mathbf{\hat O_{3D}} = \{ \hat O_{3D_i} || \hat O_{3D_i} = (\hat B_{3D_i}, \hat s_i, \hat c_i, \hat {id}_i) \}^H_{i=1}$ with $\hat B_{3D_i} = (\hat L_{3D_i}, \hat S_{3D_i})$, $\hat L_{3D_i} = (\hat x_{3D_i}, \hat y_{3D_i}, \hat z_{3D_i})$ denotes the center coordinates of the bounding box, $\hat S_{3D_i} = (\hat l_i, \hat w_i, \hat h_i)$ represents its length, width, and height, $\hat s_{3D_i}$ is the mean confidence score of the associated 2D bounding boxes, $\hat c_i$ is the class ID, $\hat {id}_i$ is the object ID and $H$ is the total number of objects currently being tracked. Occasionally, due to erroneous 2D Spatial Data Association, the same object may get different IDs in each view, causing multiple nearby 3D bounding boxes for the same object (see Figure \ref{fig:3dboxfusion1}). To address this issue, we perform weighted fusion of these bounding boxes.

The 3D bounding box fusion algorithm consists of the following steps:
\begin{enumerate}
    \item First, the list $\mathbf{\hat O_{3D}}$ is sorted in descending order based on the volume of the bounding boxes.
    \item Initialize two empty lists, $\mathbf{G}$ and $\mathbf{F}$, to store the box groups and the fused bounding boxes, respectively. Define a boolean array $\mathbf{U} = \{ U_i || U_i = False \}^H_{i=1}$ to indicate whether each box has been merged.
    \item Iterate through each bounding box $\hat O_{3D_i}$ in the list $\mathbf{\hat O_{3D}}$. If $U_i = True$, it is skipped. If $U_i = False$, it is added to $\mathbf{G}$ and the iteration is temporarily paused at index $i$.
    \item {After that, we iterate through each subsequent bounding box $\hat O_{3D_j}$ in the list $\mathbf{\hat O_{3D}}$, starting from $j = i + 1$. We compute the overlap between the two boxes using the 3D IoA, which is calculated similarly to the 2D IoA \cite{ioa}, except that volume is used instead of area. Let $\hat V_i$ denote the volume of $\hat B_{3D_i}$. The 3D IoA is computed as:
        \begin{equation}
            \text{IoA}(\hat V_i, \hat V_j) = \frac{\left|\hat V_i \cap \hat V_j \right|}{\left| \min(\hat V_i, \hat V_j) \right|}
        \end{equation}
    If IoA is greater than a threshold $thr$, $\hat O_{3D_j}$ is added to $\mathbf{G}$ and set $U_j = True$. In our experiments, we set $thr = 0.1$ for the most optimal result.
    }
    \item {After identifying the boxes with significant overlap, we perform fusion over the $T$ boxes in the list $\mathbf{G}$ $(1 \le T \le H)$. The center location and dimensions of the fused box are computed as the weighted average of the center locations and dimensions of the boxes in $\mathbf{G}$, where the weights are the volumes of the respective boxes:
        \begin{align}
            \hat L_{3D} &= \frac{\sum^T_{k=1} \hat V_k * \hat L_{3D_k}}{\sum^T_{k=1} \hat V_k} \\
            \hat S_{3D} &= \frac{\sum^T_{k=1} \hat V_k * \hat S_{3D_k}}{\sum^T_{k=1} \hat V_k}
        \end{align}
    We assign the object ID of the fused box as the minimum object ID among the boxes in $\mathbf{G}$. Thus, the confidence score and class ID of the fused box are taken from the box with the corresponding object ID. Finally, the fused bounding box $\hat O_{3D}$ is added to $\mathbf{F}$, and set $\mathbf{G}$ is empty.
    \item The fusion process is repeated until all elements in $\mathbf{U}$ are set to $True$.
    }
\end{enumerate}

\subsubsection{Yaw Refinement}

To enhance the accuracy of the 3D bounding box orientation, the Yaw Refinement approach estimates the yaw angle of each target based on its movement trajectory. Every 10 frames, we sample the world coordinates of the tracked target. Using the positions before and after the sampling interval, we calculate the yaw angle $\theta$ representing the orientation of the target in the global coordinate system. The yaw angle is computed as follows:

\begin{equation}
\theta = \arctan\left(\frac{y_{t} - y_{t-10}}{x_{t} - x_{t-10}}\right)
\end{equation}

where $(x_{t-10},y_{t-10})$ and $(x_t,y_t)$ are the coordinates of the target at frame $t-10$ and frame $t$, respectively. To ensure robust and meaningful updates, the yaw angle is only updated if the Euclidean distance between the old position and the new position exceeds a threshold of $0.15$.
\section{Experiments and Results}
\label{sec:results}

\subsection{Datasets}

The AIC25 Multi-Camera Perception Dataset \cite{Tang25AICity25} features 23 synthetic scenes with the NVIDIA Omniverse Platform. All videos in the dataset have the resolution of $1920 \times 1080$ at 30 FPS. The 19 scenes in the training and validation set feature up to 50 cameras in warehouse, hospital, and laboratory environments, while the 4 scenes in the test set feature only Warehouse scenes, though with a large variation of target classes. The AIC25 MTMC dataset provides 3D bounding box annotations with multiple object classes. Depth maps are also included as an extra modality.

\subsection{Evaluation Metrics}

The Higher Order Tracking Accuracy (HOTA) \cite{luiten2021hota} is a metric that provides a balanced evaluation of trackers. In addition to the unified score that ranks trackers' overall performances, the metric can be decomposed into the three primary aspects of the tracker: Detection Accuracy (DetA), Association Accuracy (AssA), and Localization Accuracy (LocA). In this challenge, the metric that matches ground-truth to predicted trajectories is adapted to the 3D IoU, calling for accurate estimation of the boxes' location, scale, and rotation in the 3D coordinate system.

\subsection{Implementation Details}

\noindent\textbf{Object Detection.} We sampled 2D detections from the training set at one image every 120 frames, except for the \texttt{Warehouse\_014} scene, where we sampled an image every 5 frames. For pedestrians and the two robotic classes, we utilized the pose estimation model to only sample quality detections where the head and neck points, or the two legs, are visible. The Co-DETR model is trained for 16 epochs with the SwinTransformer backbone \cite{swintransformer}.

\noindent\textbf{Re-Identification Model.} For each ID in the MTMC dataset, we took a random set of up to 15 pedestrian images per camera to create the ReID dataset. We follow the two-phase training paradigm of CLIP-ReID \cite{li2023clip}, with the learnable text tokens of the first stage being trained for 120 epochs with the learning rate of $3.5 \times 10^{-4}$, and the ViT image encoder being trained for 60 epochs with the learning rate of $5 \times10^{-6}$ with the Adam optimizer.
    
\noindent\textbf{Segmentation.} We employed the Segment Anything Model 2 \cite{sam2} with the Hiera Large backbone (SAM 2.1) for instance segmentation, leveraging its ability to generate high-quality masks from bounding box prompts. The model was initialized with pre-trained weights from the SAM 2.1 checkpoint and optimized for memory efficiency using xFormers' attention mechanism \cite{xFormers2022}. The segmentation process was parallelized across multiple processes (8 by default) to handle large-scale video data, with a batch size of 50 frames per process. For each camera frame, all bounding boxes were combined to generate a single binary mask, which was subsequently eroded using a 3x3 kernel to refine mask boundaries. 

\subsection{Results}

\subsubsection{2D Multi-Target Multi-Camera Tracking}

\begin{table}[!ht]
\centering
\centering
\resizebox{\columnwidth}{!}{%
\begin{tabular}{lcccc} 
\hline
\textbf{Methods}                                                       & \textbf{HOTA}  & \textbf{DetA}  & \textbf{AssA}  & \textbf{LocA}   \\ 
\hline
Baseline \cite{kim2024cluster}                                                               & 48.63          & 82.95          & 28.54          & 94.15           \\
\begin{tabular}[c]{@{}l@{}}+ MOT ID Consistency\\Matching\end{tabular} & 69.09          & \textbf{83.01} & 57.57          & \textbf{94.19}  \\
+ Track Splitting                                                   & \textbf{69.23} & \textbf{83.01} & \textbf{57.79} & 94.18 \\   
\hline
\end{tabular}
}
\caption{Ablation study of the proposed 2D MTMC components on the validation set.}
\label{table:ablation-2d}
\end{table}

We first present the result of our proposed 2D Multi-Camera Tracking algorithm on the validation set of the AIC25 MTMC Dataset in Table \ref{table:ablation-2d}. We utilized the same Spatial Data Association results across our experiments, while the baseline for the Temporal Data Association is the Hungarian matching method implemented in the work of Nota \cite{kim2024cluster}, the third-placed team at the 2024 AI City Challenge's MTMC track \cite{Shuo24AIC24}. It can be seen that our MOT ID Consistency matching mechanism helped improve the HOTA score by 20.46\%, with the association accuracy AssA improving by 29.03\%, demonstrating significant improvement in tracking result throughout the video, where the target's local ID information, which was generated by robust motion and appearance information, helps facilitate more accurate ID assignment. Furthermore, the track splitting mechanism based on MOT ID Consistency further boosted the association accuracy, helping to mitigate the occasional false clustering results caused by occlusions or early target entries.

\subsubsection{3D Multi-Target Multi-Camera Tracking}

\begin{table}[!ht]
\centering
\resizebox{\columnwidth}{!}{%
\begin{tabular}{lcccc} 
\hline
\textbf{Methods}     & \textbf{HOTA}  & \textbf{DetA}  & \textbf{AssA}  & \textbf{LocA}   \\ 
\hline
Baseline             & 13.24          & 13.72          & 14.71          & 39.65           \\
+ MOT ID Consistency & 18.83          & 18.44          & 21.19          & 41.25           \\
+ Late 3D Aggregation & 28.18          & 31.22          & 26.55          & 46.15           \\
+ Yaw Refinement         & \textbf{28.75} & \textbf{31.55} & \textbf{27.61} & \textbf{49.02}  \\
\hline
\end{tabular}
}
\caption{Ablation study of the proposed components.}
\label{table:ablation-3d}
\end{table}

The collective results of our proposed components on the public test set are shown in Table \ref{table:ablation-3d}. For the baseline and MOT ID Consistency methods, a fixed-size 3D bounding box is created at the ground-plane coordinate of the track for each class. In the 3D MTMC setting, the MOT ID Consistency module as a whole not only helped to increase the AssA point by 7\%, but also DetA by approximately 5\%, thanks to its aid in selecting a more representative foot point of the target for placing the 3D box. Our Late 3D Bounding Box Aggregation module represents the biggest boost of 13\% in DetA and 10\% in HOTA, emphasizing the importance of accurate 3D box localization for this dataset's overall performance. Finally, the Yaw Refinement component helped to improve the localization accuracy, pushing the final HOTA score to 28.75\%.

\begin{table}[!ht]
\centering
\begin{tabular}{cclc} 
\toprule
\textbf{Rank} & \textbf{Team ID} & \multicolumn{1}{c}{\textbf{Team Name}}                & \textbf{HOTA}   \\ 
\hline
1             & 65               & ZV                                                    & 69.91           \\
2             & 15               & SKKU-AutoLab                                          & 63.13           \\
\textbf{3}    & \textbf{133}     & \textcolor[rgb]{0.2,0.2,0.2}{\textbf{TeamQDT (Ours)}} & \textbf{28.75}  \\
4             & 116              & UTE AI Lab                                            & 25.39           \\
...           & ...              & \multicolumn{1}{c}{...}                               & ...             \\
\bottomrule
\end{tabular}
\caption{Leaderboard of Track 1 in the 2025 AI City Challenge.}
\label{table:leaderboard}
\end{table}

In the final Public Leaderboard of the 2025 AI City Challenge Track 1, our tracking pipeline achieved third place with the HOTA score of 28.75\%.
\section{Conclusions}
\label{sec:conclusions}

In this paper, we introduced a robust and flexible framework designed to extend existing online 2D Multi-Camera tracking systems into a 3D-capable tracking system without any modification. By effectively leveraging depth information, we presented a new approach that reconstructs targets into point-cloud spaces based on 2D MTMC results, computes the 3D bounding boxes through clustering techniques and refines their orientation. The proposed tracker also features an improved data association mechanism that effectively assigns global target IDs based on the consistency of the trajectories' local IDs. Our proposed system achieved 3rd place in Track 1 of the 2025 AI City Challenge \cite{Tang25AICity25}.

{
    \small
    \bibliographystyle{ieeenat_fullname}
    \bibliography{main}
}

\end{document}